Grammar construction methods for extended deterministic expressions
Xiaoying Mou and Haiming Chen
2022

Work supported by the National Natural Science Foundation of China under Grant No 61872339

# 扩展确定性正则表达式的文法构造方法

2019 年 6 月 21 日

# 目 录

# 第 1 章　背景与意义

扩展正则表达式（简称扩展表达式）以及确定性正则表达式的概念、优势、研究现状以及应用。引出关于扩展确定性正则表达式（DREs(&,#)）的研究尚且不够，希望研究此类表达式的文法表示方式，从而加深用户对此类表达式的理解，促进其实际应用。具体内容以及国内外相关工作，可参考文章 [1]，投至 Information and Computaion 的期刊 [2]，以及硕士毕业论文《扩展确定性正则表达式的判定与文法表示》。

第 2 章 预备知识# 第 2 章 预备知识

## 2.1 扩展正则表达式

假设 $\Sigma$ 是一个字母表，即一个有限非空字符集合，那么 $\Sigma$ 上的所有有限字符串集合为 $\Sigma^*$。我们用 $\varepsilon$ 代表空串，即出现 0 次符号的串。$\Sigma$ 上的标准正则表达式递归地定义如下：$\emptyset$、$\varepsilon$ 和 $a$ 是标准正则表达式；对于任意两个标准正则表达式 $E_1$ 和 $E_2$，并操作 $E_1 + E_2$，连接操作 $E_1 \cdot E_2$，以及星号操作 $E^*$ 的结果仍是标准正则表达式。实际中，我们常将表达式 $E + \varepsilon$ 或 $\varepsilon + E$ 简写为问号操作 $E?$，在不引起歧义的情况下省略书写连接符号。

扩展表达式扩展了标准正则表达式的定义，增加了无序符号 [3]：$E_1 \& E_2$，增加了计数符号 [4]：$E_1^{[m,n]}$，其中下界 $m$ 与上界 $n$ 满足条件：$m \in \mathbb{N}, n \in \mathbb{N} \setminus \{0\} \cup \{\infty\}$，$m \leqslant n$，$\mathbb{N}$ 为集合 $\{0, 1, 2, ...\}$。值得注意的是，此时 $E^* = E^{[0,\infty]}$，$E? = E^{[0,1]}$。在扩展表达式中，计数符号优先级最高，随后依次是连接，并，无序符号。

假设 $u, v$ 是两个任意字符串，我们用 $u \& v$ 表示两者任意无序交错后的字符串集合。若其中一个为空串，则 $u \& \varepsilon = \varepsilon \& u = \{u\}$。若 $u, v$ 都是非空串，我们提取出它们的前缀字符可得到 $au' = u$，$bv' = v$，则 $u \& v = \{a(u' \& v)\} \cup \{b(u \& v')\}$。该定义可以自然扩展到正则语言的定义：对于给定的两个语言 $L_1$ 与 $L_2$，$L_1 \& L_2 = \{w \in u \& v \mid u \in L_1, v \in L_2\}$ [5]。因此，对于扩展正则表达式 $E$，我们记 $L(E)$ 为 $E$ 所描述的正则语言，$L(E)$ 定义如下：

$$L(\emptyset) = \emptyset \qquad L(\varepsilon) = \{\varepsilon\} \qquad L(a) = \{a\}, a \in \Sigma$$

$$L(E_1 + E_2) = L(E_1) \cup L(E_2) \qquad L(E_1 \cdot E_2) = L(E_1) \cdot L(E_2)$$

$$L(E_1 \& E_2) = L(E_1) \& L(E_2) \qquad L(E_1^{[m,n]}) = \bigcup_{m \leqslant i \leqslant n} L(E_1)^i$$

**例 2.1.** 给定扩展表达式 $E = (ab) \& (cd + e)$，它接受的语言为 $L(E) = \{abcd, acbd, acdb, cadb, cabd, cdab, abe, aeb, eab\}$。

扩展表达式中存在以下简化规则：$E + \emptyset = \emptyset + E = E$，$E\emptyset = \emptyset E = \emptyset$，$E \& \emptyset = \emptyset \& E = \emptyset$，$\emptyset^{[m,n]} = \varepsilon$，$E + \varepsilon = \varepsilon + E = E$，$E \& \varepsilon = \varepsilon \& E = E$，$\varepsilon^{[m,n]} = \varepsilon$。对于一个简化后的表达式 $E$，要么 $E$ 中不包含 $\emptyset$，否则 $E = \emptyset$。因为我们的研究并不关注 $E = \emptyset$ 的情况，所以在之后的讨论中我们将假设所有用到的表达式都是经过简化的，不考虑存在 $\emptyset$ 的情况。

## 2.2 确定性及相关定义

对于一个扩展表达式 $E$,我们给 $E$ 中出现的每个字符 $a \in \Sigma$ 标上不同的下标 $i \in \{1, 2, ...\}$,使得每个字符 $a_i$ 仅出现一次,此时被标号的 $E$ 称为标号表达式,记作 $\overline{E}$。将一个带标号表达式 $\overline{E}$ 去掉标号的形式记作 $\overline{\overline{E}}$,即 $\overline{\overline{E}} = E$。我们很容易将加标号与去标号的操作扩展到字符串上。扩展表达式 $E$ 中出现的不同字符的集合记为 $sym(E)$,值得注意的是,当表达式是带标号的,$sym(\overline{E})$ 是 $\overline{E}$ 中所有出现的带标号的字符集合。那么 $sym(E)$ 上的所有有限字符串集合,可记为 $sym(E)^*$。

**例 2.2.** 给定 $E = (ab)\&(ad + b)$,其加标号形式可为 $\overline{E} = (a_1 b_1)\&(a_2 d_1 + b_2)$。

**定义 2.1.** [6] 给定一个表达式 $E$,如果对于任意字符串 $uxv, uyw \in L(\overline{E})$,其中 $u, v, w \in sym(\overline{E})^*$,$x, y \in sym(\overline{E})$,如果 $\overline{x} = \overline{y}$,则 $x = y$,那么我们称 $E$ 是确定性表达式。

**例 2.3.** 给定表达式 $E = (a?\&b)a$,根据确定性定义可知 $E$ 是不确定性的。因为存在字符串 $b_1 a_1 a_2, b_1 a_2 \in L((a_1?\&b_1)a_2)$,满足 $\overline{a_1} = \overline{a_2} = a$,但是 $a_1 \neq a_2$。

直观来讲,确定性要求的是字符串在匹配表达式的过程中,不需要往前看其它的字符,只需依据当前字符就能唯一确定匹配表达式中的位置。当表达式是不确定性的,则表达式中必存在两个带标号字符是 *compete*。

**定义 2.2.** [6] 给定一个标号表达式 $\overline{E}$,$\overline{E}$ 中两个带标号字符 $x, y$ 是 *compete*,当且仅当存在两个字符串 $uxv, uyw \in L(\overline{E})$,其中 $u, v, w \in sym(\overline{E})^*$。

## 2.3 表达式的其它相关定义

扩展表达式 $E$ 的大小记为 $|E|$,它等于 $E$ 中的所有字符、操作符的个数,以及所有计算符号中的上下界的二进制表示的长度总和。

对于扩展表达式 $E$,我们定义一个布尔函数 $\lambda(E)$ 来判定 $L(E)$ 是否接受空串 $\varepsilon$,如果 $\lambda(E) = true$ 则 $\varepsilon \in L(E)$,反之 $\lambda(E) = false$。函数定义如下:

$$\lambda(\varepsilon) = true \qquad \lambda(a) = false \ (a \in \Sigma)$$

$$\lambda(E_1 + E_2) = \lambda(E_1) \vee \lambda(E_2) \qquad \lambda(E_1 E_2) = \lambda(E_1) \wedge \lambda(E_2)$$

$$\lambda(E_1 \& E_2) = \lambda(E_1) \wedge \lambda(E_2) \qquad \lambda(E_1^{[m,n]}) = \begin{cases} true & if\ m = 0, \\ false, & otherwise; \end{cases}$$

**定义 2.3.** 对于每个扩展表达式 $E$，有以下集合定义：

$$first(E) = \{a \mid au \in L(E), a \in sym(E), u \in sym(E)^*\}$$

$$followlast(E) = \{a \mid uav \in L(E), u \in L(E), u \neq \varepsilon, a \in sym(E), v \in sym(E)^*\}$$

对于表达式 $E$，$first(E)$ 是 $E$ 接收的字符串中所有可能出现在首位的字符集合；在 $E$ 接收的字符串中，若存在字符串的前缀字符串非空且被 $E$ 接收，那么紧随其后的字符即属于 $followlast(E)$。

**定义 2.4.** 给定一个扩展表达式 $E$，$first$ 集合的计算方法如下：

$$first(\varepsilon) = \emptyset \qquad\qquad first(a) = \{a\},\ a \in \Sigma$$

$$first(F + G) = first(F) \cup first(G) \qquad first(F\&G) = first(F) \cup first(G)$$

$$first(F \cdot G) = \begin{cases} first(F) \cup first(G), & if\ \varepsilon \in L(F), \\ first(F), & otherwise; \end{cases}$$

$$first(F^{[m,n]}) = first(F)$$

**定义 2.5.** 给定一个标号扩展表达式 $E$，$followlast$ 集合的计算方法如下：$\overline{E} = \varepsilon$ or $x : followlast(\overline{E}) = \emptyset$

$\overline{E} = \overline{F} + \overline{G} : followlast(\overline{E}) = followlast(\overline{F}) \cup followlast(\overline{G})$

$\overline{E} = \overline{F} \cdot \overline{G}$ :

$$followlast(\overline{E}) = \begin{cases} followlast(\overline{G}) & if\ \varepsilon \notin L(\overline{G}) \\ followlast(\overline{F}) \cup first(\overline{G}) \cup followlast(\overline{G}) & if\ \varepsilon \in L(\overline{G}) \end{cases}$$

$\overline{E} = \overline{F}\&\overline{G}$ :

$$followlast(\overline{E}) = \begin{cases} followlast(\overline{F}) \cup followlast(\overline{G}) & if\ \varepsilon \notin L(\overline{F}), \varepsilon \notin L(\overline{G}) \\ followlast(\overline{F}) \cup followlast(\overline{G}) \\ \quad \cup first(\overline{G}) & if\ \varepsilon \notin L(\overline{F}), \varepsilon \in L(\overline{G}) \\ followlast(\overline{F}) \cup followlast(\overline{G}) \\ \quad \cup first(\overline{F}) & if\ \varepsilon \in L(\overline{F}), \varepsilon \notin L(\overline{G}) \\ followlast(\overline{F}) \cup followlast(\overline{G}) \\ \quad \cup first(\overline{F}) \cup first(\overline{G}) & if\ \varepsilon \in L(\overline{F}), \varepsilon \in L(\overline{G}) \end{cases}$$

$\overline{E} = \overline{F}^{[m,n]}$ :

$$followlast(\overline{E}) = \begin{cases} followlast(\overline{F}) \cup first(\overline{F}) & if\ \overline{E}.flexible \\ followlast(\overline{F}) & otherwise \end{cases}$$

其中在处理 $E = F^{[m,n]}$ 时，我们采用文章 [7] 为带计数表达式所提出的 *flexible* 属性。该属性是为了说明形如 $E = F^{[m,n]}$ 的表达式在匹配输入的字符串时，是否已经到达了计数符号的上界，如果未达到，则 $E.flexible = true$，反之 $E.flexible = false$。如下定义可扩展应用于扩展表达式，因为无序的增加并不会影响子表达式 $F$ 出现的次数，无序在这里可视为一种特殊的连接操作，所以计算 *flexible* 属性时，我们对无序符号采取与连接符号相同的操作，具体见文章 [7] 的算法 *markFlexible*()。

**定义 2.6** ([7])． 给定 $\overline{E}$ 是一个标号的带计数表达式，它的子表达式 $\overline{F} = \overline{G}^{[m,n]} (n \geq 2)$ 是 *flexible*，仅当存在字符串 $uws \in L(\overline{E})$，其中 $w \in L(\overline{F})^l$, $l \geq 1$ 并且 $w \in L(\overline{F})^{l'} L(\overline{G})^k$，其中 $l' < l$，$k < n$。

**例 2.4.** 表达式 $\overline{E_1} = (a_1^{[1,2]} + b_1)^{[2,2]}$ 是 *flexible*，因为依据定义 2.6，存在 $w = a_1 a_1 \in L(\overline{E_1})$ 且 $wb_1 \in L(\overline{E_1})$。也就是，存在字符串 $a_1 a_1 \in (\overline{E_1})$，它可匹配 $(a_1^{[2,2]})^{[1,1]}$，但未达到此表达式中计数符号的上界 2。表达式 $\overline{E_2} = (a_1^{[2,3]} + b_1)^{[2,2]}$ 不是 *flexible* 的，虽然它与 $\overline{E_1}$ 形式相近，但找不到一个未达到计数符号的上界就被它接收的字符串。

## 2.4 文法相关定义

扩展表达式类可用以下正则文法描述：

$start \rightarrow \mathbf{0} \mid \tau$

$\tau \rightarrow \varepsilon \mid a \mid (\tau \cdot \tau) \mid (\tau + \tau) \mid (\tau \& \tau) \mid (\tau)^* \quad (a \in \Sigma)$

一个上下文无关文法 $G = (N, T, P, S_0)$ 是满足如下条件的四元组：

(1) $N$ 是非终止符的集合；

(2) $T$ 是终止符的集合；

(3) $S_0 \in N$ 是开始符号；

(4) $P \subseteq N \times (N \cup T)^*$ 是一个由产生式组成的有限集合。

在文法 $G$ 中，一个非终止符 $N_0$ 能够经过推导最终得出一个字符串 $w$ 时（记作 $N_0 \stackrel{*}{\Rightarrow} w$），我们称 $N_0$ 是 *useful*。一个产生式中出现的非终止符都是 *useful*，那么这个产生式是 *valid*。一个文法 $G$ 所描述的语言我们记为 $L(G)$，它是文法从开始符号能推导出的字符串的集合 $L(G) = \{w \in T^* \mid S_0 \stackrel{*}{\Rightarrow} w\}$。当字符串 $w \in L(G)$ 时，我们称 $w$ 是 $G$ 生成的一个句子。

# 第 3 章  第一种文法构造方法

由定义 2.1可见，确定性的语义定义在标号表达式上。已有的大部分确定性判定工作也是基于标号表达式进行的 [2, 6–9]，其中文章 [2] 给出了带标号的扩展确定性表达式的特性。我们考虑基于该性质设计文法，显然，仅在假设字母表为 $\{a\}$ 时，可能的终止符就包括 $a_1, a_2, \ldots$。终止符的个数是无限的，这种文法是难以设计与实现的。因此，我们考虑去掉标号的预处理，进一步研究确定性的原表达式的性质，从而设计相应的文法表示。

## 3.1  扩展确定性表达式的特性

我们先将文章 [2] 中的定理 4 基于表达式的结构，进行重写如下:

**定理 3.1.** 给定一个扩展表达式 $E$ 是确定性的，当且仅当，

1. $\overline{E} = \varepsilon \; or \; x \; (x \in \overline{\Sigma})$ : $\overline{E}$ 是确定性的。
2. $\overline{E} = \overline{E}_1 + \overline{E}_2$ : $\overline{E}$ 是确定性的，当且仅当 $\overline{E}_1, \overline{E}_2$ 都是确定性的，且 $first(\overline{E}_1) \cap first(\overline{E}_2) = \emptyset$。
3. $\overline{E} = \overline{E}_1 \cdot \overline{E}_2$ : $\overline{E}$ 是确定性的，当且仅当 $\overline{E}_1, \overline{E}_2$ 都是确定性的，且
   - 当 $\lambda(\overline{E}_1)$，则 $first(\overline{E}_1) \cap first(\overline{E}_2) = \emptyset$ 且 $followlast(\overline{E}_1) \cap first(\overline{E}_2) = \emptyset$；
   - 当 $\neg\lambda(\overline{E}_1)$，则 $followlast(\overline{E}_1) \cap first(\overline{E}_2) = \emptyset$。
4. $\overline{E} = \overline{E}_1 \& \overline{E}_2$ : $\overline{E}$ 是确定性的，当且仅当 $\overline{E}_1, \overline{E}_2$ 都是确定性的，且 $sym(\overline{E}_1) \cap sym(\overline{E}_2) = \emptyset$。
5. $\overline{E} = \overline{E}_1^{[m,n]}$ : $\overline{E}$ 是确定性的，当且仅当 $\overline{E}_1$ 是确定性的，且当 $n > 1$，对于所有 $x \in followlast(\overline{E}_1)$，所有 $y \in first(\overline{E}_1)$，若 $\overline{x} = \overline{y}$，则 $x = y$。

带计数的表达式具有以下属性 [10]，这些属性也同样适用于扩展表达式。因为我们在 [2] 中已证明了为扩展表达式所提出的 $first, followlast$ 集的计算规则，都符合集合的定义2.3。

**引理 3.2.** 对于一个扩展表达式 $E$,

  a. $sym(E) = \overline{sym(\overline{E})}$。
  b. $first(E) = \overline{first(\overline{E})}$。
  c. $followlast(E) \supseteq \overline{followlast(\overline{E})}$；当 $E$ 是确定性的, $followlast(E) = \overline{followlast(\overline{E})}$。

借助引理 3.2，我们可以将定理 3.1 中 (1)-(4) 的标号操作去除。由引理 3.2 中的 c 可知，定理 3.1 声明 (5) 中的标号不能直接去除。声明 (5) 本意是在 $n > 1$ 时，确保不会有两个不相同的标号符号 $x, y$，去掉标号后相同，而且满足 $x \in followlast(\overline{E_1})$，$y \in first(\overline{E_1})$。若直接去掉标号，就成为要求在 $n > 1$ 的时，$followlast(E_1) \cap first(E_1) = \emptyset$，显然这样改写会比声明 (5) 的限制更强，因为它连标号符号相同的情况都不允许出现。于是，为得到声明 (5) 在原表达式上的等价操作，我们定义以下布尔函数 $\mathcal{W}$。

**定义 3.1.** 布尔函数 $\mathcal{W}(E)$ 在扩展表达式上定义如下：

$$\mathcal{W}(\varepsilon) = \mathcal{W}(a) = true$$

$$\mathcal{W}(E_1 + E_2) = \mathcal{W}(E_1) \wedge \mathcal{W}(E_2) \wedge \big(followlast(E_1) \cap first(E_2) = \emptyset\big)$$
$$\wedge \big(followlast(E_2) \cap first(E_1) = \emptyset\big)$$

$$\mathcal{W}(E_1 E_2) = \big(followlast(E_2) \cap first(E_1) = \emptyset\big) \wedge \Big(\big(\neg\lambda(E_1) \wedge \neg\lambda(E_2)\big)$$
$$\vee \big(\lambda(E_1) \wedge \neg\lambda(E_2) \wedge \mathcal{W}(E_2)\big) \vee \big(\lambda(E_1) \wedge \lambda(E_2) \wedge \mathcal{W}(E_1) \wedge \mathcal{W}(E_2)\big)$$
$$\vee \big(\neg\lambda(E_1) \wedge \lambda(E_2) \wedge \mathcal{W}(E_1) \wedge \big(first(E_1) \cap first(E_2) = \emptyset\big)\big)\Big)$$

$$\mathcal{W}(E_1 \& E_2) = \mathcal{W}(E_1) \wedge \mathcal{W}(E_2)$$

$$\mathcal{W}(E_1^{[m,n]}) = \mathcal{W}(E_1)$$

**引理 3.3.** 对于一个确定性扩展表达式 $E$，如果 $\mathcal{W}(E) = true$ 当且仅当对于所有 $x \in followlast(\overline{E})$，所有 $y \in first(\overline{E})$，若 $\overline{x} = \overline{y}$，则 $x = y$。

**证明.** 我们仍是依据 $E$ 的结构进行归纳证明，这里仅给出 $E = E_1 \& E_2$ 的情况的详细证明。其它的情况在文章 [10] 中可见。因为 $E$ 是确定性的，由定义 2.1 与定理 3.1 可知，$E_1$ 与 $E_2$ 都是确定性的。

($\Rightarrow$)：因为 $\mathcal{W}(E) = true$，$\mathcal{W}(E_1) = \mathcal{W}(E_2) = true$。取 $x \in followlast(\overline{E})$，$y \in first(\overline{E})$，且 $\overline{x} = \overline{y}$。当 $\varepsilon \notin L(E_1)$ 且 $\varepsilon \notin L(E_2)$，$first(E) = first(E_1) \cup first(E_2)$，$followlast(E) = followlast(E_1) \cup followlast(E_2)$。假设 $x \in followlast(E_1)$，$y \in first(E_1)$（或者 $x \in followlast(E_2)$，$y \in first(E_2)$）。因为 $\mathcal{W}(E_1) = true$，$\mathcal{W}(E_2) = true$，以及 $E_1, E_2$ 都是确定性的，根据归纳假设有 $x = y$。假设 $x \in followlast(E_1)$，$y \in first(E_2)$（或者 $x \in followlast(E_2)$，$y \in first(E_1)$）。因为 $E$ 是确定性的，由定理 3.1 与引理 3.2 得到 $sym(E_1) \cap sym(E_2) = \emptyset$。因此有 $x = y$。对于情况 $\varepsilon \notin L(E_1)$，$\varepsilon \in L(E_2)$ 或 $\varepsilon \in L(E_1)$，$\varepsilon \notin L(E_2)$ 或者 $\varepsilon \in L(E_1)$，$\varepsilon \in L(E_2)$ 时，同理可证。 ($\Leftarrow$)：基于定义 2.4 与定义 2.5，$first(E) = first(E_1) \cup first(E_2)$，$followlast(E) \supseteq followlast(E_1) \cup followlast(E_2)$。那么右侧已成立的条件，可等价表述为对于所有 $x \in followlast(\overline{E_1})$ 且所有 $y \in first(\overline{E_1})$，或者所有 $x \in followlast(\overline{E_2})$ 且所有 $y \in first(\overline{E_2})$，若 $\overline{x} = \overline{y}$，则 $x = y$。通过归纳假设，可得到 $\mathcal{W}(E_1) = true$，$\mathcal{W}(E_2) = true$。因此，$\mathcal{W}(E) = \mathcal{W}(E_1) \land \mathcal{W}(E_2) = true$。 □

**推论 3.4.** 对于表达式 $E = E_1^{[m,n]}$，$E$ 是确定性的，当且仅当 $E_1$ 是确定性的且 $\mathcal{W}(E_1) = true$。

证明. 由定理 3.1 与引理 3.3 可证。 □

于是，我们可以得到扩展确定性表达式基于原表达式的直接特性，如下：

**定理 3.5.** 对于一个扩展表达式 $E$：

1. $E = \varepsilon \text{ or } a$：$E$ 是确定性的。
2. $E = E_1 + E_2$：$E$ 是确定性的，当且仅当 $E_1, E_2$ 都是确定性的，且 $first(E_1) \cap first(E_2) = \emptyset$。
3. $E = E_1 \cdot E_2$：$E$ 是确定性的，当且仅当 $E_1, E_2$ 都是确定性的，且
   - 当 $\lambda(E_1)$，$first(E_1) \cap first(E_2) = \emptyset$ 且 $followlast(E_1) \cap first(E_2) = \emptyset$；
   - 当 $\neg \lambda(E_1)$，$followlast(E_1) \cap first(E_2) = \emptyset$。
4. $E = E_1 \& E_2$：$E$ 是确定性的，当且仅当 $E_1, E_2$ 都是确定性的，且 $sym(E_1) \cap sym(E_2) = \emptyset$。
5. $E = E_1^{[m,n]}$：$E$ 是确定性的，当且仅当 $E_1$ 是确定性的，且若 $n > 1$，$\mathcal{W}(E_1) = true$。

证明. 基于引理 3.2 与推理 3.4，可将定理 3.1 等价改写为此条定理。 □

## 3.2 文法表示

本节将给出扩展确定性表达式的文法表示。我们先将定理 4 转化为推导系统 $\mathcal{T}$，模拟一个扩展确定性表达式的推导生成过程，从而设计文法。

### 3.2.1 推导系统

在推导系统中，$\vdash r$ 表明一个表达式 $r$ 满足确定性。一条推导 $\dfrac{\vdash r_1 \ldots \vdash r_n \quad c_1 \ldots c_m}{\vdash r}$ 表示：如果 $\vdash r_1 \ldots \vdash r_n$ 满足确定性且条件 $c_1 \ldots c_m$ 成立，那么 $r$ 也满足确定性。我们称 $r$ 是可推导的，当且仅当存在一棵推导树满足 $r$ 是根节点 [11]。

推导系统 $\mathcal{T}$ 包含以下推导规则，每一条规则对应定理 4中的声明。

Base: $\dfrac{}{\vdash \varepsilon \vdash a\ (a \in \Sigma)}$   Union: $\dfrac{\vdash r \ \vdash s \ \ first(r) \cap first(s) = \emptyset}{\vdash r + s}$   Inter: $\dfrac{\vdash r \ \vdash s \ \ sym(r) \cap sym(s) = \emptyset}{\vdash r \& s}$

Concat: $\dfrac{\vdash r \ \vdash s \ \ followlast(r) \cap first(s) = \neg\lambda(r) \vee first(r) \cap first(s) = \emptyset}{\vdash rs}$   Count: $\dfrac{\vdash r \ \mathcal{W}(r)}{\vdash r^{[m,n]}}$

**定理 3.6.** 一个扩展表达式 $r$ 是确定性的，当且仅当 $r$ 在 $\mathcal{T}$ 上是可推导的。

证明. 给定一个扩展确定性表达式 $r$，构建 $r$ 的语法树则会与其推导树同构，因此 $r$ 在 $\mathcal{T}$ 上是可推导的。对于每个从 $\mathcal{T}$ 推导出的 $r$，$r$ 是确定性的，因为 $\mathcal{T}$ 中的每个规则对应于定理 4中的每条声明。 □

### 3.2.2 DREs(&, #) 的文法

从 $\mathcal{T}$ 中可以看出，要构造一个扩展确定性表达式，我们必须保证它的所有子表达式都是确定性的，而且在每种操作符下，其子表达式的 $first$, $followlast$, $sym$ 集合和函数 $\lambda$, $\mathcal{W}$ 符合定理 4中的条件。以 Union: $r + s$ 为例，我们可以给出任意确定性表达式 $r'$ 来替换 $r$，只要满足条件 $first(r') \cap first(s) = \emptyset$，那么 $r' + s$ 仍然是确定性的。

我们设计文法 $G_{idre} = (N, T, P, S_0)$ 来描述 DREs(&, #)。让 $\Sigma = \{a_1, ..., a_n\}$，那么终止符为 $T = \{+, \cdot, \&, (,), ?, *, [,], m, n\} \cup \Sigma$。非终止符为 $N = \mathbb{R} \cup \{R_0, A, B, C\}$，每个 $\mathbb{R}$ 中的非终止符为 $R^{F,L,S,\alpha,\beta}$，其中 $F, L, S \subseteq \Sigma$，$\alpha, \beta \in \{0, 1\}$ (1 等价于 $true$，0 等价于 $false$)。$R^{F,L,S,\alpha,\beta}$ 用于描述一组表达式 $r \in \text{DREs}(\&, \#)$ 满足: $F = first(r), L = followlast(r), S = sym(r), \alpha = \lambda(r), \beta = \mathcal{W}(r)$。我们令所有非终止符都是开始符 $S_0 = N$。

关于产生式 $P$，我们将参照 $\mathcal{T}$ 中的推导规则，因为每条针对不同操作符的规则类都对应于一组表达式。同时我们在其中增加了 $first$, $followlast$, $sym$, $\lambda$ 和 $\mathcal{W}$ 的计算规则。产生式 $P$ 如下所示:

Base :   $R^{\{a_i\},\emptyset,\{a_i\},0,1} \to a_i, i \in \{1, 2, \cdots\}$    $R_0 \to \emptyset \mid \varepsilon$

Union :   $R^{F,L,S,\alpha,\beta} \to \bigcup\limits_{con1} (R^{F_1,L_1,S_1,\alpha_1,\beta_1} + R^{F_2,L_2,S_2,\alpha_2,\beta_2})$

Concat :   $R^{F,L,S,\alpha,\beta} \to \bigcup\limits_{con2} (R^{F_1,L_1,S_1,\alpha_1,\beta_1} \cdot R^{F_2,L_2,S_2,\alpha_2,\beta_2})$

Inter :   $R^{F,L,S,\alpha,\beta} \to \bigcup\limits_{con3} (R^{F_1,L_1,S_1,\alpha_1,\beta_1} \& R^{F_2,L_2,S_2,\alpha_2,\beta_2})$

Opt :   $R^{F,L,S,1,\beta} \to (R^{F,L,S,\alpha,\beta})?$    Star : $R^{F,L,S,1,1} \to \bigcup\limits_{L=F_1 \cup L_1} (R^{F,L_1,S,\alpha,1})*$

Count :   $R^{F,L,S,1,1} \to \bigcup\limits_{L=F_1 \cup L_1} (R^{F,L_1,S,\alpha,1})[C]$    $C \to nAm$   $A \to nAm \mid B$   $B \to nB \mid n$

其中:

$$con1 \stackrel{def}{=} F = F_1 \cup F_2, L = L_1 \cup L_2, S = S_1 \cup S_2, \alpha = \alpha_1 \vee \alpha_2, F_1 \cap F_2 = \emptyset,$$
$$\beta = (\beta_1 \wedge \beta_2 \wedge (F_1 \cap L_2 = \emptyset) \wedge (L_1 \cap F_2 = \emptyset)).$$
$$con2 \stackrel{def}{=} (\alpha_1 \wedge (F = F_1 \cup F_2) \wedge (F_1 \cap F_2 = \emptyset)) \vee (\neg \alpha_1 \wedge (F = F_1)), S = S_1 \cup S_2, \alpha = \alpha_1 \wedge \alpha_2,$$
$$(\alpha_2 \wedge (L = L_1 \cup L_2 \cup F_2)) \vee (\neg \alpha_2 \wedge (L = L_2)), L_1 \cap F_2 = \emptyset, \beta = (F_1 \cap L_2 = \emptyset) \wedge$$
$$((\neg \alpha_1 \wedge \neg \alpha_2) \vee (\alpha_1 \wedge \neg \alpha_2 \wedge \beta_2) \vee (\alpha_1 \wedge \alpha_2 \wedge \beta_1 \wedge \beta_2) \vee (\neg \alpha_1 \wedge \alpha_2 \wedge \beta_1 \wedge F_1 \cap F_2 = \emptyset)).$$
$$con3 \stackrel{def}{=} F = F_1 \cup F_2, S = S_1 \cup S_2, \alpha = \alpha_1 \wedge \alpha_2, \beta = \beta_1 \wedge \beta_2, S_1 \cap S_2 = \emptyset,$$
$$(\neg \alpha_1 \wedge \neg \alpha_2 \wedge (L = L_1 \cup L_2)) \vee (\neg \alpha_1 \wedge \alpha_2 \wedge (L = L_1 \cup L_2 \cup F_2)) \vee$$
$$(\alpha_1 \wedge \neg \alpha_2 \wedge (L = L_1 \cup L_2 \cup F_1)) \vee (\alpha_1 \wedge \alpha_2 \wedge (L = L_1 \cup L_2 \cup F_1 \cup F_2)).$$

我们使用符号 (⊍) 来代表一组有相同左侧非终止符的产生式。当字母表 $\Sigma$ 固定时，则 $T$，$N$、$P$ 都是有限的。显然，上面定义的 $G_{idre}$ 是上下文无关文法。与标准确定性表达式的文法相比，我们将非终止符的参数组扩增了 $S$，并增加了几组处理无序与计数符号的产生式。

值得注意的是，在将推导规则 $\mathcal{T}$ 中的 **Count**：$r^{[m,n]}$ 情况转化为文法的产生式类时，共分了 3 类 **Opt**、**Star** 与 **Count** 来体现。对于 $r^{[m,n]}$，我们需要慎重考虑 $m$ 与 $n$ 的取值，因为 $m,n$ 的取值将影响到表达式 $r^{[m,n]}$ 是否具有 *flexible* 属性，而该属性又将影响的 *followlast* 集合（也就是非终止符五元组中的 $L$）的计算。它们要满足的基本条件为：$m \leq n$，而且 $m \in \mathbb{N}, n \in \mathbb{N} \setminus \{0\} \cup \{\infty\}$。为了避免引入复杂的计算公式来求解 *flexible* 属性（参照文章 [7]，至少需要在文法中额外引入的一个 double 类型的综合属性与一个 int 类型的继承属性才能求解当前表达式是否为 *flexible* 的），在这里，生成文法时我们先只考虑 $m < n$ 的情况。在此情况下，当 $m = 0, n = 1$ 时，我们简写为 $r?$，此时由定义 2.5 知 $followlast(r?) = followlast(F)$，即产生式中的 **Opt** 类；当 $m > 0, n > 1$ 时由定义 2.6 可得，$r^{[m,n]}$ 是 *flexible* 的，那么由定义 2.5 知 $followlast(F^{[m,n]}) = followlast(F) \cup first(F)$。为了能够生成两个正整数 $m,n$，且满足 $n > m, n > 1, m > 0$，我们用非终止符 $C$ 来描述语言 $L = \{n^n m^m \mid n > m, n > 1, m > 0\}$。因此我们只需要统计生成的字符串中 $n,m$ 的次数，即可得到计数符号的上界 $n$ 与下界 $m$，这就是产生式中的 **Count** 类。因为 $n$ 的出现次数是有限的，所以又增加了 **Star** 类，使得 $n$ 可以取值 $\infty$。

**定理 3.7.** $DREs(\&, \#)$ 可由上下文无关文法 $G_{idre}$ 描述。

证明. 我们将证明：（1）所有 $G_{idre}$ 能推导出的字符串都是 $DREs(\&, \#)$，可通过对推导的长度进行归纳；（2）所有 $DREs(\&, \#)$ 都可以由 $G_{idre}$ 推导得到，可通过对表达式的结构进行归纳。

**(1)**：在文法 $G_{idre}$ 中，如果存在一个非终止符 $R^{F,L,S,\alpha,\beta} \stackrel{+}{\Rightarrow} r$，那么 $r$ 是确定

性的且满足 $first(r) = F, followlast(r) = L, sym(r) = S, \lambda(r) = \alpha, \mathcal{W}(r) = \beta$。

**基本**：如果推导只有一步，根据 $G_{idre}$，那么 $r = a$ 或 $\varepsilon$，其中 $a \in \Sigma$。$r$ 肯定是确定性的。

**归纳步骤**：假设推导有 $k + 1$ 步，且前 $k$ 步满足该定理，也就是 $G_{idre}$ 中存在相应的非终止符产生确定性表达式。如果第 $(k+1)$ 步推导用到 Inter 类的产生式：$R^{F,L,S,\alpha,\beta} \Rightarrow R_1 \& R_2 \overset{*}{\Rightarrow} r$。因为 $R_1$ 或 $R_2$ 的推导不超过 $k$ 步，那么有 $R_1 \overset{*}{\Rightarrow} r_1$，$R_2 \overset{*}{\Rightarrow} r_2$，其中 $r_1, r_2$ 是确定性表达式。则 $r = r_1 \& r_2$。依据产生式规则 Inter，$r, r_1$ 与 $r_2$ 满足其中的 $con3$，于是有 $sym(r_1) \cap sym(r_2) = \emptyset$。又因为 $r_1, r_2$ 是确定性的，根据定理 4 得到 $r$ 是确定性的。而且存在 $F = first(r), L = followlast(r), S = sym(r), \alpha = \lambda(r), \beta = \mathcal{W}(r)$，其中由 $con3$ 知他们的取值，例如：$F = first(r_1) \cup first(r_2), \alpha = \lambda(r_1) \wedge \lambda(r_2) \cdots$。如果第 $(k+1)$ 步推导用到其它类的产生式规则，同理可证。

**(2)**：如果存在一个扩展确定性表达式 $r$，以及五个元素 $F = first(r), L = followlast(r), S = sym(r), \alpha = \lambda(r), \beta = \mathcal{W}(r)$，那么在文法 $G_{idre}$ 中将有一个 *useful* 非终止符 $R^{F,L,S,\alpha,\beta}$，满足 $R^{F,L,S,\alpha,\beta} \overset{*}{\Rightarrow} r$。

$r = a(\varepsilon)$ 时，参照文法 $G_{idre}$ 的 **Base** 类产生式规则，有 $R^{\{a\},\emptyset,\{a\},0,1}(R_{\emptyset})$ 满足条件。

$r = r_1 \& r_2$：因 $r$ 是确定性的，则由定理 4 知 $r_1, r_2$ 是确定性的。由归纳假设，可得出在文法 $G_{idre}$ 中存在 *useful* 非终止符 $R_1$ 与 $R_2$，满足 $R_1^{F_1,L_1,S_1,\alpha_1,\beta_1} \overset{*}{\Rightarrow} r_1$ 且 $R_2^{F_2,L_2,S_2,\alpha_2,\beta_2} \overset{*}{\Rightarrow} r_2$，其中 $F_1 = first(r_1), L_1 = followlast(r_1), S_1 = sym(r_1), \alpha_1 = \lambda(r_1), \beta_1 = \mathcal{W}(r_1)$ and $F_2 = first(r_2), L_2 = followlast(r_2), S_2 = sym(r_2), \alpha_2 = \lambda(r_2), \beta_2 = \mathcal{W}(r_2)$。依据 $first, followlast, sym, \lambda$ 与 $\mathcal{W}$ 的计算规则，我们可以得到，$first(r) = first(r_1) \cup first(r_2), (\neg\lambda(r_1) \wedge \neg\lambda(r_2) \wedge followlast(r) = followlast(r_1) \cup followlast(r_2)) \vee (\neg\lambda(r_1) \wedge \lambda(r_2) \wedge followlast(r) = followlast(r_1) \cup followlast(r_2) \cup first(r_2)) \vee (\lambda(r_1) \wedge \neg\lambda(r_2) \wedge followlast(r) = followlast(r_1) \cup followlast(r_2) \cup first(r_1)) \vee (\lambda(r_1) \wedge \lambda(r_2) \wedge followlast(r) = followlast(r_1) \cup followlast(r_2) \cup first(r_1) \cup first(r_2)), sym(r) = sym(r_1) \cup sym(r_2)$ and $\lambda(r) = \lambda(r_1) \wedge \lambda(r_2), \beta(r) = \beta(r_1) \wedge \beta(r_2)$。这符合 $G_{idre}$ 中 Inter 类产生式的规则 $con3$，因此存在 $R^{F,L,S,\alpha,\beta} \Rightarrow R_1 \& R_2 \overset{*}{\Rightarrow} r_1 \& r_2$，也就是 $R^{F,L,S,\alpha,\beta} \overset{*}{\Rightarrow} r$。

其它情况 ($r = r_1 + r_2, r = r_1 \cdot r_2, r = r_1?, r = r_1^*, r = r_1^{[m,n]}(m < n, n > 1, m > 0)$) 可同理证明。 □

**定理 3.8.** $DREs(\&, \#)$ 不能被正则文法描述。

证明. 文章 [1] 证明了标准确定性正则表达式不能由正则文法描述。因为标准确定性正则表达式不像正则语言那样在同态下保存封闭性。因此，IDRE 作为标准确定性正则表达式的超集类，故而也不能由正则文法定义。 □

### 3.2.3 文法的优化

我们分析文法 $G_{idre}$ 的规模，即其中非终止符集 $N$ 的大小与产生式 $P$ 的条数。对于给定的字母表 $\Sigma$，先分析 $N$ 的数目，因为 $first, followlast$ 与 $sym$ 集合各可能有 $2^{|\Sigma|}$ 个取值，$N$ 的大小，记为 $|N|$ 约为 $2^{3|\Sigma|+2} + 2$。每一条产生式最多会有 3 个非终止符出现，那么产生式 $P$ 的条数，记为 $|P|$，约为 $\mathcal{O}(2^{9|\Sigma|})$。显然 $|N|$ 与 $|P|$ 的数量级将随着 $|\Sigma|$ 的增大而爆发，这使得从中找到 useful 的非终止符与 valid 的产生式的任务极其艰巨。注：因为在 $G_{idre}$ 文法中，**Count** 类产生式会产生无限种可能性的 $m$ 与 $n$ 的情况，且除了需要生成的计数符号不同外它与 **Star** 类的规则完全相同，我们在统计 $|N|$ 时不计入 $C, A, B$ 三个非终止符，在统计 $|P|$ 时，不计入 **Count** 类的产生式。

我们通过观察文法 $G_{idre}$ 中产生式规则的约束，发现了 useful 非终止符的性质，从而得到如下的引理。引理中的优化规则能够加速化简文法，去除无效产生式。因为只有产生式中所有的非终止符都是 useful 的，该条产生式才是 valid。

**引理 3.9.** 文法 $G_{idre}$ 中的非终止符 $R^{F,L,S,\alpha,\beta}$ 是 useful 的，当且仅当它同时满足以下四个条件: (1) $S \cap F \neq \emptyset$; (2) $F \cup L \subseteq S$; (3) $(F \cap L = \emptyset) \rightarrow (\beta = 1)$; (4) $(\alpha = 0$ and $\beta = 1) \rightarrow (F \not\subseteq L)$。

证明. ($\Rightarrow$) (1) $S$ 代表着 $sym$ 集合，即包括所有表达式中出现的字母表，$F$ 是表达式的 $first$ 集合，由定义 2.4 知，除了表达式是 $\varepsilon$ 时，为 $\{\varepsilon\}$，其余情况下为字符的集合。因为我们将 $\varepsilon$ 单独用非终止符 $R_\varepsilon$ 推导得出，其余情况下并不生成 $\varepsilon$ 而是用产生式类 **Opt** 替代，所以 $S$ 与 $F$ 都不为空，$S \cap F \neq \emptyset$。(2) $F$ 与 $L$ 都是 $S$ 的子集，所以 $F \cup L \subseteq S$。(3) 当 $F \cap L = \emptyset$，不存在字符 $a, b$ 满足 $a = b$，$a \in F, b \in L$，由引理 3.3 得到 $W = true$ ($\beta = 1$)。(4) 当 $\alpha = 0$ 且 $\beta = 1$ 时，假设 $F \subseteq L$，那么至少存在字符 $s \in F \cap L$。因为 $\alpha = 0$，$R^{F,L,S,\alpha,\beta}$ 从产生式 **Base**, **Union**, **Concat**, 或 **Inter** 推导而来的。如果，它来自 **Opt**, **Star** 或 **Count**，那么 $\alpha = 1$ 与前提条件矛盾。若是 **Base**，我们可得到 $L = \emptyset$，这与 $s \in L$ 相矛盾。若是 **Union**，$R^{F,L,S,\alpha,\beta} \rightarrow R^{F_1,L_1,S_1,\alpha_1,\beta_1} + R^{F_2,L_2,S_2,\alpha_2,\beta_2}$，参照文法 $G_{idre}$ 的 con1，我们得出 $F = F_1 \cup F_2, L = L_1 \cup L_2$，那么 $s$ 可能存在于 $F_1 \cap L_1, F_1 \cap L_2, F_2 \cap L_1$ 或 $F_2 \cap L_2$

中，由定义 3.1 知，这些情况都将导致 $\mathcal{W} = false\,(\beta = 0)$，这与前提条件 $\beta = 1$ 相违背。对于 **Concat, Inter** 的情况，同理可得。

($\Leftarrow$) 我们将证明当以上四个条件同时成立时，存在一个扩展确定性表达式 $r$ 使得 $first(r) = F, followlast(r) = L, sym(r) = S, \lambda(r) = \alpha, \mathcal{W}(r) = \beta$。也就意味着 $R^{F,L,S,\alpha,\beta}$ 是 *useful*。

当 $F \cap L = \emptyset$ 且满足其它条件时，我们假设 $F = \{a_1, \ldots, a_m\}, L = \{b_0, \ldots, b_p\}$，其中 $a_i\,(i \in [1, m]), b_j\,(j \in [0, p])$ 是不同的字符，那么 $S = \{a_1, \ldots, a_m, b_0, \ldots, b_p\}$。因为满足条件（3），则有 $\beta = 1$。因此我们考虑以下几种情况：

- 情况 1: 当 $\alpha = 1$ 时，我们可以构建表达式 $r = ((a_1 + \ldots + a_m) \cdot (b_0^* + \ldots + b_p^*))?$，它满足 $first(r) = F, followlast(r) = L, sym(r) = S, \alpha = 1, \beta = 1$。

- 情况 2: 当 $\alpha = 0$，因为我们保证了 $F \nsubseteq L$，所以表达式 $r = (a_1 + \cdots + a_m) \cdots (b_0^* + \ldots + b_p^*)$ 可以满足 $first(r) = F, followlast(r) = L, sym(r) = S, \alpha = 0, \beta = 1$。

当 $F \cap L \neq \emptyset$ 且满足其它条件时，我们假设 $F = \{a_0, \ldots, a_m, c_1, \ldots, c_n\}$ and $L = \{b_0, \ldots, b_p, c_1, \ldots, c_n\}$，那么 $S = \{a_0, \ldots, a_m, b_0, \ldots, b_p, c_1, \ldots, c_n\}$。我们考虑以下几种情况：

- 情况 1: 当 $\alpha = 0, \beta = 0$ 时，我们可以构建表达式 $r = (a_0 + \ldots + a_m + c_1 \ldots + c_n) \cdot (b_0^* + \ldots + b_p^* + c_1^* \ldots + c_n^*)$，它满足 $first(r) = F, followlast(r) = L, sym(r) = S, \alpha = 0, \beta = 0$。

- 情况 2: 当 $\alpha = 1, \beta = 0$ 时，表达式 $r = ((a_0 + \ldots + a_m + c_1 + \ldots + c_n) \cdot (b_0^* + \ldots + b_p^* + c_1^* \ldots + c_n^*))?$ 满足 $first(r) = F, followlast(r) = L, sym(r) = S, \alpha = 1, \beta = 0$。

- 情况 3: 当 $\alpha = 1, \beta = 1$ 时，若 $m \neq 0$，我们可以构建表达式 $r = (((a_1 + \ldots + a_m) \& (c_1^* + \ldots + c_n^*)) \cdot (b_0^* + \ldots + b_p^*))?$；若 $m = 0$，表达式 $r = ((c_1 \& \ldots \& c_n) \cdot (b_0^* + \ldots + b_p^*))^*$ 满足 $first(r) = F, followlast(r) = L, sym(r) = S, \alpha = 1, \beta = 1$。

- 情况 4: 当 $\alpha = 0, \beta = 1$ 时，因为满足条件（4），$F \nsubseteq L$，所以必然存在一个字符 $a \in F$ 但是 $a \notin L$。因此我们将 $F$ 原本的赋值改为 $F = \{a_1, \ldots, a_m, c_1, \ldots, c_n\}$。表达式 $r = ((a_1 + \ldots + a_m) \& (c_1^* + \ldots + c_n^*)) \cdot (b_0^* + \ldots + b_p^*)$ 可以满足 $first(r) = F, followlast(r) = L, sym(r) = S, \alpha = 0, \beta = 1$。

□

## 3.3 实验分析

在本节中,我们将在小字母表上实现文法 $G_{idre}$。一方面我们将通过展示文法的非终结符与产生式大小,来表明引理 3.9中优化规则的有效性。另一方面我们将对比我们简化的文法与标准确定性表达式的文法 $s\text{-}G_{dre}$ [1] 的规模。所有实验在一台 2.2 GHz Intel Core i7 和 16G RAM 的计算机上进行。

我们可以利用引理 3.9简化文法 $G_{idre}$,并将简化后的文法表示为 $s\text{-}G_{idre}$,它只包含 *useful* 的非终止符和 *valid* 的产生式。我们在小字母表上实现 $G_{idre}$ 与 $s\text{-}G_{idre}$,它们的规模见表 3.1。其中 $|N|$ 是非终止符集合的大小,$|P|$ 是产生式的数量,$|\Sigma|$ 是字母表的大小。

**表 3.1 文法 $G_{idre}$ 与化简后文法 $s\text{-}G_{idre}$ 的规模对比**

| $|\Sigma|$ | $G_{idre}$ | | $s\text{-}G_{idre}$ | | $\frac{|P| \ in \ s\text{-}G_{idre}}{|P| \ in \ G_{idre}}(100\%)$ |
|---|---|---|---|---|---|
| | $|N|$ | $|P|$ | $|N|$ | $|P|$ | |
| 1 | 33 | 2,097 | 6 | 15 | 0.668% |
| 2 | 257 | 103,810 | 44 | 1,116 | 1.074% |
| 3 | 2,049 | 4,926,467 | 282 | 46,865 | 0.951% |
| 4 | 16,385 | – | 1,652 | 1,495,482 | – |

从表 3.1可知,优化规则是非常有效的,尤其对于减少生产数量,使其可降低大约两个数量级。当 $n = 3$ 时,$G_{idre}$ 中约 99.05% 的产生式不是 *valid* 的。当 $n = 4$ 时,生成 $G_{idre}$ 的产生式时就会因执行时间过长而无法记录,这也表明了优化规则的必要性。

我们也实现了标准确定性表达式的文法 $s\text{-}G_{dre}$ [1],与 $s\text{-}G_{idre}$ 做对比,实验结果见下图 4.1。左侧是 $|N|$ 的大小,右侧是 $|P|$ 的大小,字母表大小取值 1 ~ 4。

我们可以发现,无论从非终止符还是产生式来看,文法 $s\text{-}G_{idre}$ 与 $s\text{-}G_{dre}$ 的规模都随之字母表的增多而大幅度增长,其中以 $s\text{-}G_{idre}$ 中产生式数目的增长最为剧烈。文法 $s\text{-}G_{idre}$ 比 $s\text{-}G_{dre}$ 规模增加的部分,主要是由无序符号的增加引起的,因为我们之前已提到不将无限的 **Count** 类产生式计算在内。当 $|\Sigma| = 4$ 时,$s\text{-}G_{idre}$ 中产生式的数目已到达约 149 万条,可见该文法的规模是庞大的,会使其实用性受限。

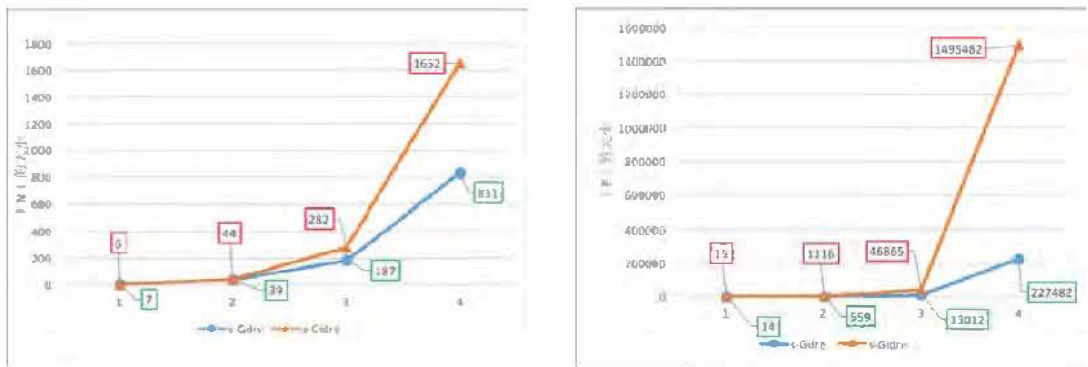

图 3.1 文法 $s\text{-}G_{idre}$ 与 $s\text{-}G_{dre}$ 的规模对比

# 第 4 章   第二种文法构造方法

我们观察定理 4可得，只有形如 $F^{[m,n]}$ 的表达式才会调用函数 $\mathcal{W}(F)$，也就是表达式 $F$ 可以连续地迭代多次。满足此类要求的表达式被称为具有属性 continuing [10]，记作 $ct(F) = true$。其形式化定义如下：

**定义 4.1** ([10]). $F$ 是表达式 $E$ 的一个子表达式，$F$ 具有 continuing 属性 (i.e., $ct(F) = true$)，仅当对于任意字符串 $w_1, w_2 \in L(F)$，存在 $u, v \in (sym(E) \cup b)^*$ 满足 $ubw_1bbw_2bv \in L(E_{F_b})$，其中 $E_{F_b}$ 是将 $E$ 中的 $F$ 替换为 $bFb$ ($b \notin sym(E)$)。

我们设计函数 $markRable$ 来计算表达式是否具有continuing 属性，见图 4.1，其中参数 $E$ 是输入的表达式，布尔参数 $N$ 将赋值于当前子表达式。函数 $markRable$ 的计算过程实际上是自上而下地遍历表达式 $E$ 的语法树。而，属性值 $N$ 则是由上一层子表达式传递给下一层子表达式。只有在两种情况下，属性值 $N$ 才会发生改变：一种是当 $E = E_1^*$ 时，则 $ct(E_1) = true = N$；另一种是当 $E = E_1 \cdot E_2$ 且 $ct(E) = true$ 时，则 $ct(E_1) = \lambda(E_2) = N$，$ct(E_2) = \lambda(E_1) = N$。

```
Function markRable(E: Expression, N: Boolean):
    case E = a or ε:  ct(E) = false;
    case E = E₁ + E₂ or E = E₁&E₂:  ct(E) = N;
            markRable(E₁, N); markRable(E₂, N);
    case E = E₁ · E₂:  ct(E) = N;
            if ct(E), then {markRable(E₁, λ(E₂)); markRable(E₂, λ(E₁));}
            else { markRable(E₁, N); markRable(E₂, N);}
    case E = E₁^[m,n]:
            if n = 1, then  ct(E) = N; markRable(E₁, N);
            else   ct(E) = N; markRable(E₁, true);
```

**图 4.1** 调用 $markRable(E, false)$ 计算 continuing 属性

于是我们可以将定理中的 $\mathcal{W}$ 函数用 $ct$ 函数替代，改写为以下定理

**定理 4.1.** 对于一个扩展表达式 $E$：

1. $E = \varepsilon$ or $a$: $E$ 是确定性的。
2. $E = E_1 + E_2$: $E$ 是确定性的，当且仅当 $E_1, E_2$ 都是确定性的，$first(E_1) \cap first(E_2) = \emptyset$，并且当 $ct(E)$ 时，$\big(followlast(E_1) \cap first(E_2) = \emptyset, followlast(E_2) \cap first(E_1) = \emptyset\big)$。

3. $E = E_1 \cdot E_2$: $E$ 是确定性的,当且仅当 $E_1, E_2$ 都是确定性的,$followlast(E_1) \cap first(E_2) = \emptyset$,并且

   - 当 $\lambda(E_1)$ 时,$first(E_1) \cap first(E_2) = \emptyset$;
   - 当 $ct(E)$ 时,$followlast(E_2) \cap first(E_1) = \emptyset$,且当 $\neg\lambda(E_1)$,$\lambda(E_2)$ 时,$\big(first(E_1) \cap first(E_2) = \emptyset\big)$。

4. $E = E_1 \& E_2$: $E$ 是确定性的,当且仅当 $E_1, E_2$ 都是确定性的,且 $sym(E_1) \cap sym(E_2) = \emptyset$。

5. $E = E_1^{[m,n]}$: $E$ 是确定性的,当且仅当 $E_1$ 是确定性的。

证明. 基于定理 4 与函数 $\mathcal{W}$ 与 $ct$ 的计算规则,可得到此条定理。 □

## 4.1 文法表示

我们将定理 4.1 转化为推导系统 $\mathcal{T}'$,模拟一个扩展确定性表达式的推导生成过程,从而设计文法。

### 4.1.1 推导系统

推导系统 $\mathcal{T}'$ 包含以下推导规则,每一条规则对应定理 4.1 中的每条声明。

**Base:** $\dfrac{}{\vdash \varepsilon \vdash a \, (a \in \Sigma)}$

**Union:** $\dfrac{\vdash r \quad \vdash s \quad first(r) \cap first(s) = \emptyset \wedge \big(\neg ct(r+s) \vee (followlast(r) \cap first(s) = \emptyset) \wedge (followlast(s) \cap first(r) = \emptyset)\big)}{\vdash r+s}$

**Inter:** $\dfrac{\vdash r \quad \vdash s \quad sym(r) \cap sym(s) = \emptyset}{\vdash r \& s}$

**Concat:** $\dfrac{\vdash r \quad \vdash s \quad (followlast(r) \cap first(s) = \emptyset) \wedge (\neg\lambda(r) \vee (first(r) \cap first(s) = \emptyset)) \wedge \big(\neg ct(r \cdot s) \vee ((followlast(s) \cap first(r) = \emptyset) \wedge ((first(r) \cap first(s) = \emptyset) \vee \lambda(r) \vee \neg\lambda(s)))\big)}{\vdash r \cdot s}$

**Count:** $\dfrac{\vdash r}{\vdash r^{[m,n]}}$

证明同上一种方法。

### 4.1.2 DREs(&, #) 的文法

根据 $\mathcal{T}'$,我们设计文法 $G'_{idre} = (N, T, P, S_0)$ 来描述 DREs(&, #)。让 $\Sigma = \{a_1, ..., a_n\}$,那么终止符为 $T = \{+, \cdot, \&, (,), ?, *, [,], m, n\} \cup \Sigma$。非终止符为 $N = \mathbb{R} \cup \{R_0, A, B, C\}$,每个 $\mathbb{R}$ 中的非终止符为 $R^{F,L,S,\alpha,\theta}$,其中 $F, L, S \subseteq \Sigma$,$\alpha, \theta \in \{0, 1\}$ (1 等价于 $true$, 0 等价于 $false$)。$R^{F,L,S,\alpha,\theta}$ 用于描述一组表达式 $r \in \mathrm{DREs}(\&, \#)$ 满足:$F = first(r), L = followlast(r), S = sym(r), \alpha = \lambda(r), \theta = ct(r)$。我们令所有非终止符都是开始符 $S_0 = N$。

关于产生式 $P$，我们将参照 $\mathcal{T}'$ 中的推导规则，因为每条针对不同操作符的规则类都对应于一组表达式。同时我们在其中增加了 $first$, $followlast$, $sym$, $\lambda$ 和 $ct$ 的计算规则。产生式 $P$ 如下所示：

$$\text{Base}: R^{\{a_i\},\emptyset,\{a_i\},\bullet,\theta} \to a_i, i \in \{1,2,\cdots\} \qquad R_\emptyset \to \emptyset \mid \varepsilon$$

$$\text{Union}: R^{F,L,S,\alpha,\theta} \to \bigcup_{con1}(R^{F_1,L_1,S_1,\alpha_1,\theta} + R^{F_2,L_2,S_2,\alpha_2,\theta})$$

$$\text{Concat}: R^{F,L,S,\alpha,\theta} \to \bigcup_{con2}(R^{F_1,L_1,S_1,\alpha_1,\theta_1} \cdot R^{F_2,L_2,S_2,\alpha_2,\theta_2})$$

$$\text{Inter}: R^{F,L,S,\alpha,\theta} \to \bigcup_{con3}(R^{F_1,L_1,S_1,\alpha_1,\theta} \,\&\, R^{F_2,L_2,S_2,\alpha_2,\theta})$$

$$\text{Opt}: R^{F,L,S,1,\theta} \to (R^{F,L,S,\alpha_1,\theta})? \qquad \text{Star}: R^{F,L,S,1,\theta} \to \bigcup_{L=F_1 \cup L_1}(R^{F,L_1,S,\alpha_1,1})*$$

$$\text{Count}: R^{F,L,S,1,\theta} \to \bigcup_{L=F_1 \cup L_1}(R^{F,L_1,S,\alpha_1,1})[C] \qquad C \to nAm \quad A \to nAm \mid B \quad B \to nB \mid n$$

其中：

$con1 \stackrel{def}{=} F = F_1 \cup F_2, L = L_1 \cup L_2, S = S_1 \cup S_2, \alpha = \alpha_1 \vee \alpha_2, F_1 \cap F_2 = \emptyset,$
$\neg\theta \vee ((F_1 \cap L_2 = \emptyset) \wedge (L_1 \cap F_2 = \emptyset)).$

$con2 \stackrel{def}{=} (\alpha_1 \wedge (F = F_1 \cup F_2) \wedge (F_1 \cap F_2 = \emptyset)) \vee (\neg\alpha_1 \wedge (F = F_1)), S = S_1 \cup S_2,$
$\alpha = \alpha_1 \wedge \alpha_2, (\alpha_2 \wedge (L = L_1 \cup L_2 \cup F_2)) \vee (\neg\alpha_2 \wedge (L = L_2)), L_1 \cap F_2 = \emptyset,$
$(\theta \wedge (\theta_1 = \alpha_2) \wedge (\theta_2 = \alpha_1)) \vee (\neg\theta \wedge (\theta = \theta_1 = \theta_2)),$
$\neg\theta((F_1 \cap L_2 = \emptyset) \wedge (\alpha_1 \wedge \neg\alpha_2 \wedge F_1 \cap F_2 = \emptyset), (\neg\alpha_1 \wedge \alpha_2 \wedge \theta_1 \wedge F_1 \cap F_2 = \emptyset)).$

$con3 \stackrel{def}{=} F = F_1 \cup F_2, S = S_1 \cup S_2, \alpha = \alpha_1 \wedge \alpha_2, \theta = \theta_1 \wedge \theta_2, S_1 \cap S_2 = \emptyset,$
$(\neg\alpha_1 \wedge \neg\alpha_2 \wedge (L = L_1 \cup L_2)) \vee (\neg\alpha_1 \wedge \alpha_2 \wedge (L = L_1 \cup L_2 \cup F_2)) \vee$
$(\alpha_1 \wedge \neg\alpha_2 \wedge (L = L_1 \cup L_2 \cup F_1)) \vee (\alpha_1 \wedge \alpha_2 \wedge (L = L_1 \cup L_2 \cup F_1 \cup F_2)).$

我们使用符号 $(\bigcup)$ 来代表一组有相同左侧非终止符的产生式。当字母表 $\Sigma$ 固定时，则 $T$，$N$，$P$ 都是有限的。显然，上面定义的 $G'_{idre}$ 是上下文无关文法。与标准确定性表达式的文法相比，我们将非终止符的参数组扩增了 $S$，并增加了几组处理无序与计数符号的产生式。

**定理 4.2.** $DREs(\&,\#)$ 可由上下文无关文法 $G'_{idre}$ 描述。(同上种方法可证)

### 4.1.3 $G_{idre}$ 与 $G'_{idre}$ 的对比

1. $G_{idre}$ 有优化规则，可以直接得到有用的非终止符，进而得到有效的产生式。因此 $G_{idre}$ 可以直接用文章 [1] 中的句子生成算法，快速随机生成扩展确定性正则表达式。$G'_{idre}$ 没有优化规则，故而只能从所有的产生式中遍历逐步删去无效产生式（这将耗时巨大，因为所有可能的产生式个数约为 $O(2^{9|\Sigma|})$），也不能应用以上的句子生成算法。

2. 我们对比文法 $s$-$G_{idre}$ 与 $s$-$G'_{idre}$（删除无用非终止符与无效的产生式的文

法 $G'_{idre}$) 的规模。在小字母表上的规模见表 4.1, 其中 $|N|$ 是非终止符集合的大小, $|P|$ 是产生式的数量, $|\Sigma|$ 是字母表的大小。由表可见两者的数量级相同, 但后者的规模略小。

表 4.1 简化后的文法 $G_{idre}$ 与 $G'_{idre}$ 的规模对比

| $|\Sigma|$ | $G_{idre}$ | | $G'_{idre}$ | |
|---|---|---|---|---|
| | $|N|$ | $|P|$ | $|N|$ | $|P|$ |
| 1 | 8 | 19 | 6 | 15 |
| 2 | 47 | 1,100 | 44 | 1,116 |
| 3 | 255 | 40,754 | 282 | 46,865 |
| 4 | 1,367 | 1,182,718 | 1,652 | 1,495,482 |

# 第 5 章　基于原表达式的确定性判定算法

正如 [1] 中，基于定理 3.1 可以得到一个基于标号扩展表达式的确定性判定算法，其算法复杂度为 $\mathcal{O}(|\Sigma||E|)$。同样地，我们可以将定理 4.1 转化为基于原表达式的 $\mathcal{O}(|\Sigma||E|)$ 确定性判定算法。伪代码见算法 1，其中省略了 $first, followlast, sym$ 集合的计算。

```
Algorithm 1: Determ_unmarked (E: Expression)
  input  : an expression E in IREs
  output : true if the expression E is deterministic or false otherwise
1  if E = a or ε then return true
2  if E = E₁ | E₂ then
3     if Determ_unmarked(E₁) and Determ_unmarked(E₂) then
4        if first(E₁) ∩ first(E₂) ≠ ∅ then
5           return false
6        if ct(E) and (followlast(E₁) ∩ first(E₂) ≠ ∅ or
              followlast(E₂) ∩ first(E₁) ≠ ∅) then
7           return false
8        else return true
9     else return false
10 if E = E₁ · E₂ then
11    if Determ_unmarked(E₁) and Determ_unmarked(E₂) then
12       if followlast(E₁) ∩ first(E₂) ≠ ∅ or (λ(E₁) and first(E₁) ∩ first(E₂) ≠ ∅)
         then
13          return false
14       if ct(E) and (followlast(E₂) ∩ first(E₁) ≠ ∅ or (first(E₁) ∩ first(E₂) ≠ ∅
              and ¬λ(E₁) and λ(E₂))) then return false
15       else return true
16    else return false
17 if E = E₁ & E₂ then
18    if Determ_unmarked(E₁) and Determ_unmarked(E₂) then
19       return sym(E₁) ∩ sym(E₂) = ∅
20    else return false
21 if E = E₁^[m,n] then return Determ_unmarked(E₁)
```

算法 1 $Determ\_unmarked$ 与 [1] 中基于标号表达式的判定算法相比，虽然时间复杂度相同（同 [1] 的分析），但是实际中，$Determ\_unmarked$ 具有 local nondeterminism-locating 的特性。该特性是指算法能够高效地、准确地局部定位到非确定性的子表达式，而不需要遍历整个表达式。我们已表达式 $E = ((a \cdot a?\&b) \cdot (c + d)?)*$ 为例，执行算法 $Determ\_unmarked(E)$，并将关键计算过程呈现在 $E$ 的语法树上，见图 5.1。我们也将 [1] 中算法 $FollowlastDer(E)$ 的执行过程记录在

语法树上，进行对比。

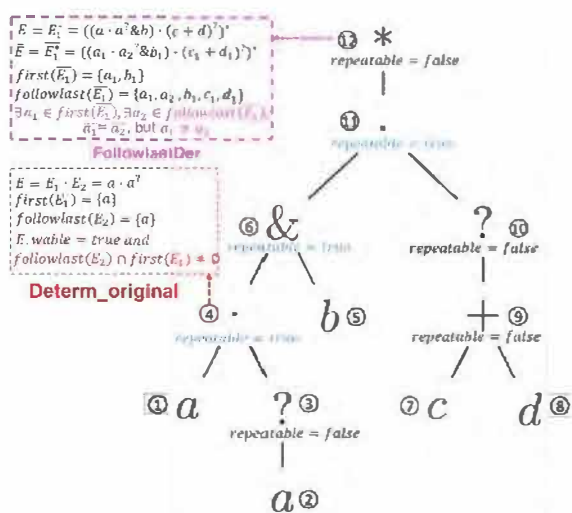

**图 5.1 表达式 $E$ 的语法树**

如图 5.1所见，算法 $Determ\_unmarked(E)$ 与 $FollowlastDer(E)$ 的返回结果都是 $E$ 是非确定性的。对于 $Determ\_unmarked(E)$，它的判定过程将依次遍历语法树的结点 $1 \to 2 \to 3 \to 4$，并且在结点 4 处，它符合算法中第 15 行的条件，返回 $false$。但是对于 $FollowlastDer(E)$，判定过程将依次遍历结点 $1 \to 2 \to \cdots \to 11 \to 12$，直到结点 12 才返回 $false$。最终，$Determ\_unmarked(E)$ 能够定位到子表达式 $a \cdot a?$ 导致了非确定性，但 $FollowlastDer(E)$ 只能定位到 $((a \cdot a?\&b) \cdot (c + d)?)^*$。显然，对于此例，$Determ\_unmarked(E)$ 的定位非确定性因素的效率更高且更准确。以此推广，我们可以得到结论，当一类非确定性表达式的非确定性因素是由子表达式 $F$ 导致的，并且 $F$ 具有 contiuing 属性时，$Determ\_unmarked(E)$ 能够发挥其 local nondeterminism-locating 特性的优势。

# 第 6 章　应用句子生成算法

我们在第三章与第四章、分别用两种方法为扩展确定性正则表达式构造了文法 $G_{idre}$ 与 $G'_{idre}$。在各章实验部分分别展示了文法简化后的规模，我们可以看到在小字母表上，文法的规模也是巨大的而且是指数级别的剧烈增长。因此在应用此类大规模文法生成句子时，需要选取合适的句子生成算法。已有的句子生成算法 [12–15] 都需要对文法遍历一遍进行预处理，这一操作不适用于 $G_{idre}$ 与 $G'_{idre}$ 这类大规模文法。我们选用文章 [1] 中的随机生成算法 Random Generation Algorithm(简称 *Ran_Generator*)，该算法不需要预处理文法即可进行随机生成，但是由于该算法只接受简化后的文法，即只包含有用非终止符与有效产生式的文法，所以 $G'_{idre}$ 不能直接应用，文法 $G_{idre}$ 的简化形式 $s\text{-}G_{idre}$ 可以。我们结合 $s\text{-}G_{idre}$ 的情况，对 [1] 中的算法 *Ran_Generator* 进行了复现，并且记录了实验结果。*Ran_Generator* 的输入是允许的字母表大小 $|\Sigma|$ 以及允许的表达式的最大长度 $l$，输出为随机生成的扩展确定性表达式 $E$，其中 $sym(E) \subseteq \Sigma, |E| \leq l$。

我们设置 $l = 30$，$|\Sigma|$ 取值从 1 到 7，记录成功随机生成一个扩展确定性表达式的耗时情况，见下表。

表 6.1 *Ran_Generator* 作用于 $s\text{-}G_{idre}$ 生成一个表达式的耗时情况

| $|\Sigma|$ | 1 | 2 | 3 | 4 | 5 | 6 | 7 |
|---|---|---|---|---|---|---|---|
| Time(s) | 0.01 | 0.12 | 0.2 | 3 | 13 | 114 | 892 |

# 参考文献